\documentclass[letterpaper, 10pt, conference]{ieeeconf} 

\IEEEoverridecommandlockouts                    
                                                          
\usepackage{multirow} 
 
\usepackage{algpseudocode}

\usepackage{rotating}
\usepackage{color}
\usepackage{subfigure}
\usepackage[linesnumbered,ruled]{algorithm2e}
\usepackage[justification=centering]{caption}
\usepackage{braket}
\usepackage{amssymb}

\definecolor{darkgreen}{rgb}{0,0.8,0}

\title{A Deep Incremental Boltzmann Machine for Modeling Context in Robots}

\author{Fethiye Irmak Do\u{g}an$^{1}$, Hande \c{C}elikkanat$^{2}$, and Sinan Kalkan$^{1}$
\thanks{$^{1}$KOVAN Research Lab, Dept. of Computer Engineering, Middle East Technical University, Ankara, Turkey
        {\tt\small \{irmak.dogan, skalkan\}@metu.edu.tr}
   $^{2}$Dept. of Computer Science and HIIT, University of Helsinki, Finland
        {\tt\small hande.celikkanat@helsinki.fi}
        }
}

\begin{document}

\maketitle
\thispagestyle{empty}
\pagestyle{empty}

\begin{abstract}

Context is an essential capability for robots that are to be as adaptive as possible in challenging environments. Although there are many context modeling efforts, they assume a fixed structure and number of contexts. In this paper, we propose an incremental deep model that extends Restricted Boltzmann Machines. Our model gets one scene at a time, and gradually extends the contextual model when necessary, either by adding a new context or a new context layer to form a hierarchy. We show on a scene classification benchmark that our method converges to a good estimate of the contexts of the scenes, and performs better or on-par on several tasks compared to other incremental models or non-incremental models. 

\end{abstract}

\section{Introduction}

Context is essential for many critical cognitive capabilities such as perception, reasoning, communication and action \cite{yeh2006situated,barsalou2009simulation}. Context helps these processes in resolving ambiguities, rectifying mispredictions, filtering irrelevant details, and adapting planning.

It is known that contexts are hierarchical structures \cite{mccarthy1993notes} such that we can think of sub-contexts of contexts. E.g., in a kitchen context, one can talk about the dishwasher context or making breakfast context that contain sub-groups of relevant objects and actions related to the kitchen context.

Robots, which are expected to share the same complex environments that we live in, should depend on  context like we do. A robot should adapt its routine tasks, e.g., when there are children around,  when it is carrying a hot drink, or when everyone is at sleep. To be able to accomplish that, a robot should be able to learn new contexts and change its behavior according to the current context. 

\begin{figure}
\centerline{
	\includegraphics[width=0.5\textwidth]{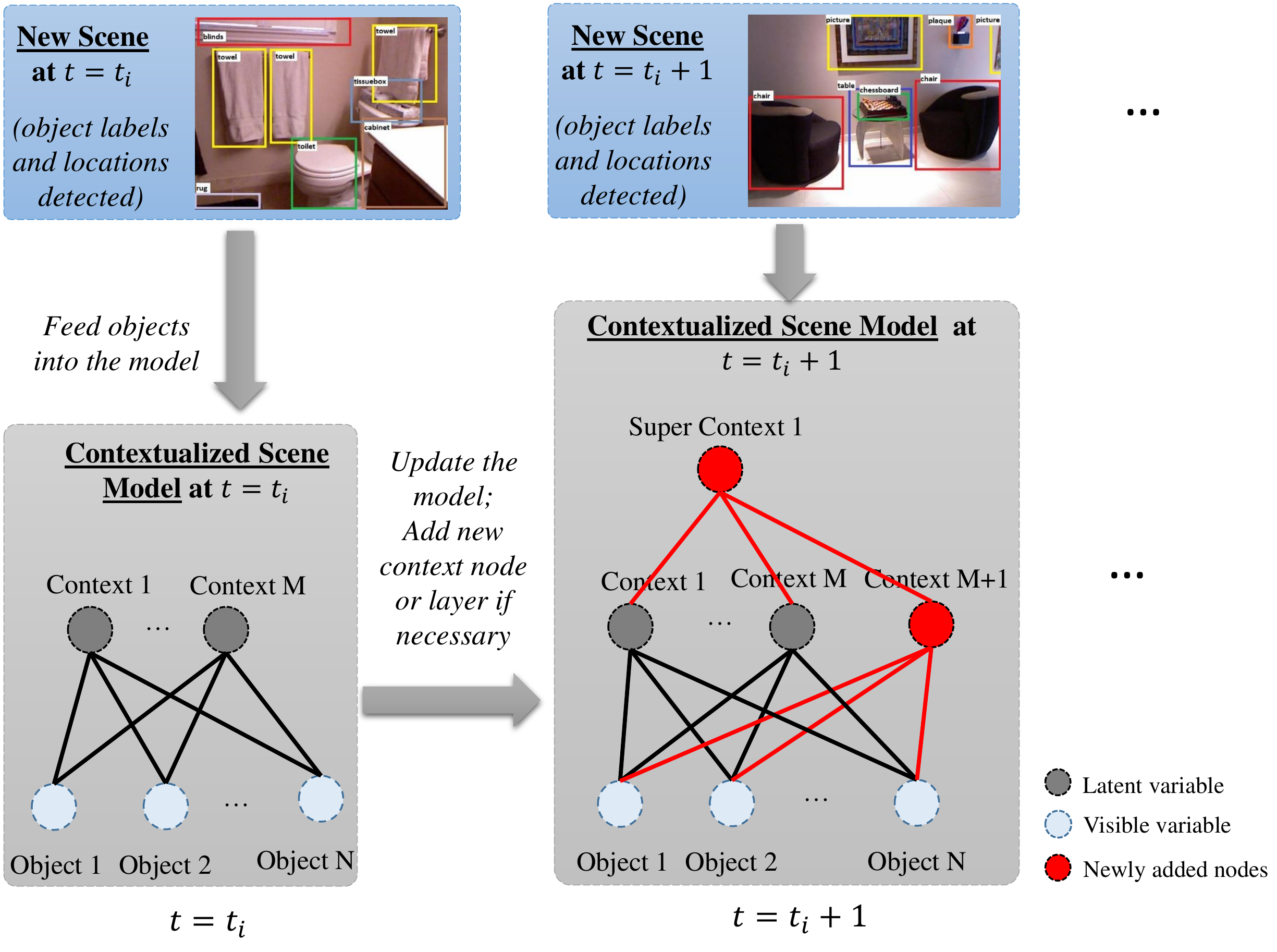}
}
\caption{An overview of the proposed system. Our model receives scenes one at a time, and updates its structure by adding a new context node or a context layer combining existing contexts if necessary.\label{fig:overview}}
\end{figure}

Learning contexts should take an incremental approach since one cannot enumerate all spatial, temporal and social configurations (situations) that can be taken as contexts. Therefore, with every experience, looking at certain signals coming from the environment or the robot, the robot should be able to update its context model.

In this work, we take an incremental approach to modeling context in robots, as shown in Fig. \ref{fig:overview}.  Although there have been many studies in incremental topic/context modeling in linguistics \cite{canini2009online} and robotics \cite{ortiz2014incremental,CelikkanatContext2014}, they are not hierarchical. There are promising hierarchical topic modeling efforts \cite{TehEtAl2006,paisley2015nested}, which however either assume a fixed depth structure or availability of all data at the model construction phase. Our approach, on the other hand, makes no assumption on the depth or the availability of the training data, and looking at its confidence in contextual representation of objects, determines when to add a new context, or a context layer to form a hierarchy.

\subsection{Related Work}
 
\textbf{Context Modeling:} In AI, McCarthy \cite{mccarthy1993notes} was known to be the first to define what context is and is not with a modeling perspective. McCarthy's definitions and formulations were in propositional logic, which was followed by similar attempts using predicate logic or  description logics \cite{buvae1993propositional,klarman2013description}. Such definitions rely on formulating a context in terms of rigid rules and relations between entities, which are difficult to enumerate in practice. 

In computer vision and pattern recognition, on the other hand, models integrated context into many problems such as object recognition \cite{torralba2003contextual,torralba2003placeandobject}, activity recognition \cite{marszalek2009actions} using probabilistic graphical models, such as Markov Random Field, Conditional Random Field, or Bayesian Networks. In these models, contextual information was provided mostly through local interactions between predictions.

In natural language processing, many models (e.g., Hidden Markov Models) have been proposed that incorporated latent variables to model \textit{hidden} information in the data. These models were followed by newer approaches such as Latent Semantic Analysis \cite{dumais1988using}, Latent Dirichlet Allocation \cite{griffiths2004finding} that have been widely used for modeling \textit{topics} (i.e., contexts) of documents, and recently scenes \cite{CelikkanatContext2014}.

There are several studies in robotics that integrate context into various robot problems, examples including \cite{jiang2012placingobjects}, which used context in determining where to place a new object in the scene; \cite{anand2013contextually}, which modeled local interactions between objects (as context) in determining their labels; and, \cite{CelikkanatContext2014}, which proposed using context in modulating object detections in a scene and planning. Our model differs from all these studies by being incremental and hierarchical.

It should also be noted that a contextual analysis is different from an ordinary clustering task in that, in clustering, elements are allocated into clusters in a 1-to-1 fashion whereas, in contextual analysis, an item can belong to one or more contexts, all at the same time, or to differing contexts at different times.

\textbf{Incremental or Hierarchical Context Modeling:}
There are incremental context or topic modeling efforts in text modeling \cite{canini2009online}, computer vision \cite{yu2015incremental} and in robotics \cite{ortiz2014incremental,CelikkanatContext2014}. These methods look at the errors or the entropy (perplexity) of the system to determine when to increment. Moreover, they are not hierarchical. There are also other methods such as Hierarchical Dirichlet Processes \cite{TehEtAl2006} or its nested version \cite{paisley2015nested} that assume the availability of all data to estimate the number of topics or assume infinite number of topics, which are both unrealistic for robots continually interacting with the environment and getting into new contexts through their lifetime.

\subsection{Contributions}
Compared to the existing studies (as briefly reviewed in the previous section), our paper makes the following major contributions:
\begin{itemize}
\item An incremental hierarchical (deep) Boltzmann Machine (BM) has been proposed, which, with each arriving new scene, determines to add a new hidden neuron or a hidden layer without making an assumption about the data or the structure.
\item We introduce two novel measures to make BM incremental and hierarchical. One measure mainly captures how strongly a neuron is represented by a hidden neuron in the next layer. This forces a context (hidden neuron) to have at least one object that strongly activates it. The second measure directs how contexts should be combined under a new upper layer in the hierarchy.
\end{itemize}

We compare our method against Restricted Boltzmann Machines (RBM) \cite{salakhutdinov2007restricted}, incremental RBM \cite{yu2015incremental}, incremental LDA \cite{CelikkanatContext2014}, Deep Boltzmann Machines (DBM) \cite{salakhutdinov2009deep} and show that it performs better in several aspects in scene modeling tasks.

\section{Background: General, Deep and Restricted Boltzmann Machines}

\begin{figure} 
\centerline{
\includegraphics[width=0.49\textwidth]{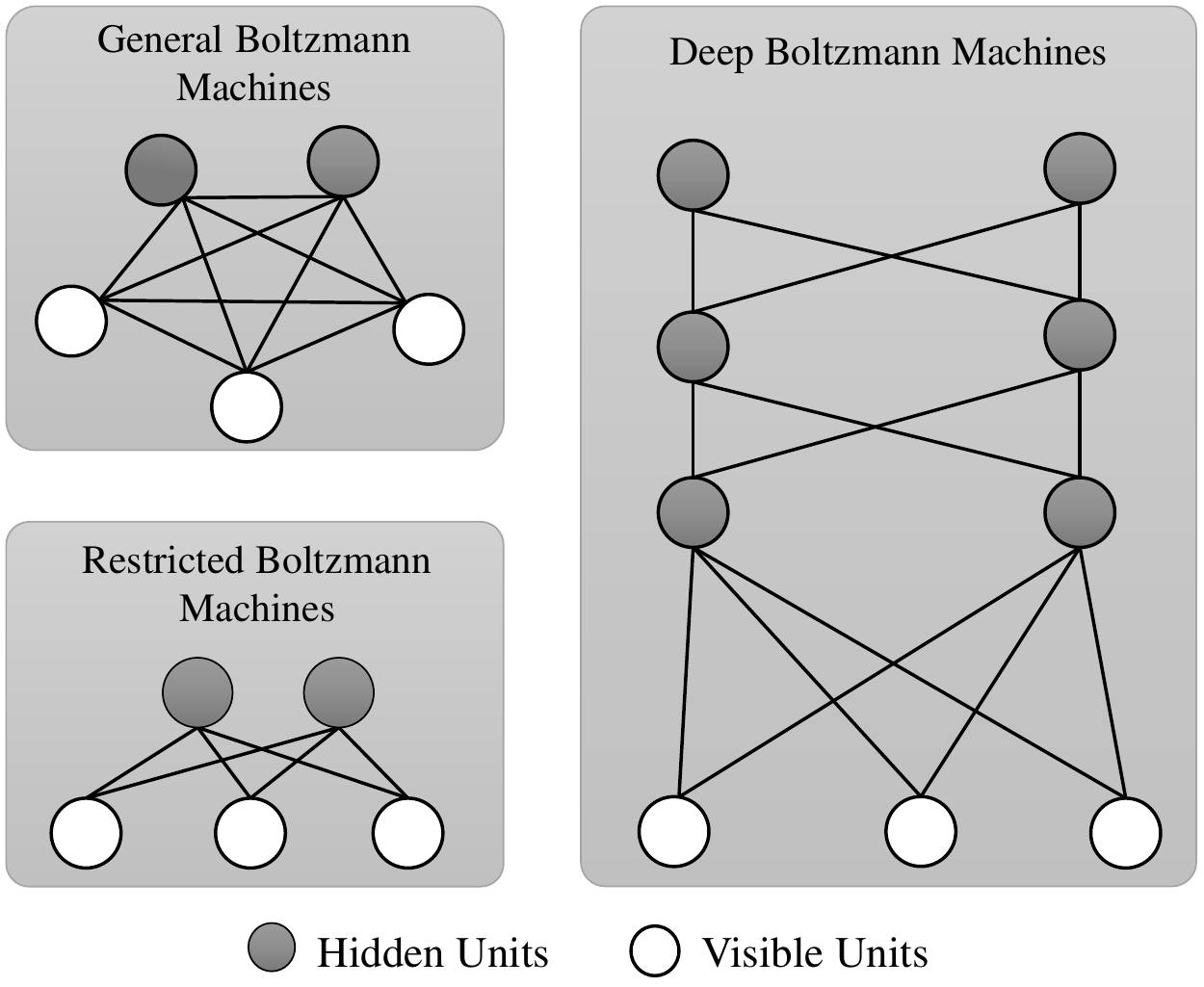}
}
\caption{A schematic comparison of Boltzmann Machines, Restricted Boltzmann Machines and Deep Boltzmann Machines. \label{fig:all_bm}}
\end{figure}
 
A Boltzmann Machine (BM) \cite{ackley1985learning} is a stochastic network composed of visible nodes $\mathbf{v}=\{v_i\}_{i=1}^{V} \subset \{0,1\}^V$ and hidden nodes $\mathbf{h}=\{h_i\}_{i=1}^{H} \subset \{0,1\}^H$ -- see also Fig. \ref{fig:all_bm}. Visible nodes and hidden nodes are connected to each other with symmetrical edges with weights $W=\{w_{ij}\}$ with $w_{ij}\in \mathbb{R}$. In general BM, there is no restriction on connections, and a node is connected to all other nodes, which, however, makes the learning and the inference problems more challenging and slow. To overcome these limitations, a restricted version of BM (called Restricted Boltzmann Machines (RBM) \cite{salakhutdinov2007restricted} or Harmonium Networks \cite{smolensky1986information}) has been proposed. Alternatively, as in Deep Boltzmann Machines (DBM) \cite{salakhutdinov2009deep}, one can form layers of hidden nodes to estimate a more reliable latent model of the data -- see Fig. \ref{fig:all_bm} for a schematic comparison.

Training an RBM consists of two phases \cite{salakhutdinov2007restricted}: (i) Positive phase: where data is clamped to the visible units $\mathbf{v}$, hidden units $\mathbf{h}^0$ are activated, and average joint activations $<v_i h_j>^0$ are calculated. (ii) Negative phase: Visible units, call it $\mathbf{v}^1$, are reconstructed from $\mathbf{h}^0$, and hidden units, $\mathbf{h}^1$, are re-estimated from $\mathbf{v}^1$. From this iteration, average joint activations $<v_i h_j>^1$ are re-calculated. 

Each weight is then updated by using these joints activations, as follows: 
\begin{equation}
    w_{ij} \leftarrow w_{ij} + \epsilon \times ( <v_ih_j>^0 - <v_ih_j>^1 )
\end{equation}

\section{Our Model: A Deep Incremental Boltzmann Machine (diBM)}

In this section, we first describe how we make one layer of incremental RBM (iRBM) and then present deep incremental BM (diBM).

\subsection{Incremental Restricted Boltzmann Machines (iRBM)}

Our first contribution is a new way to make RBM incremental. Unlike previous work which uses entropy of the system \cite{ortiz2014incremental} or the reconstruction error to make an update decision \cite{yu2015incremental}, our approach relies on calculating a confidence measure for each visible unit $v$:
\begin{equation}
  c_v \leftarrow \max\limits_{j} w_{vj},
\end{equation}
{\noindent}which essentially links a visible node's confidence to how strong it is connected to hidden neurons; if the maximum weight to hidden neurons is low, then the network has not found a suitable strong topic for that visible node yet.

Similarly, we can also define a \textit{baseline} confidence $c^{|\mathbf{h}|}_m$ for the whole model with current hidden neurons $\mathbf{h}$, using a softmax function to have a smoother behavior:
\begin{equation}
  c^{|\mathbf{h}|}_m \leftarrow \frac{1}{Z_0} \exp\left(\min\limits_{v} c_v\right),
\end{equation}
{\noindent}with $Z_0 \leftarrow \sum_v \exp(c_v)$ being the partition function. When the model is fed with new scenes ($\textbf{v}$), over time, the model will slowly fall short in representing $p(\textbf{v})$, and the model's current confidence ($c_m^{curr}\leftarrow {1/Z_0} \exp\left(\min_{v} c_v\right)$) will slightly drift away from its baseline confidence $c^{|\mathbf{h}|}_m$. When that happens, a new hidden neuron should be added to increase the model's capacity. This condition can be formulated as:
\begin{equation}
 c_m^{curr} < t^{iRBM} \times c^{|\mathbf{h}|}_m,
\label{eqn:condition_iRBM}
\end{equation}
{\noindent}where $t^{iRBM}$ is a scaling factor, controlling the system's patience (empirically set to $\exp(-0.5)$). Note that one can simplify Equation \ref{eqn:condition_iRBM} by removing $Z_0$s on both sides. 

The new neuron's weights are initialized as follows:
\begin{equation}
 w_{ik} \leftarrow \left(\sum_{j=1}^{|\textbf{h}|-1} w^{ij}\right)^{-1},
\end{equation}
{\noindent}which assigns $v_i$'s weight to $h_k$ inversely to the sum of its weights to other hidden neurons; if this sum is large, $v_i$ is strongly represented by these hidden neurons, and its weight $w_{ik}$ to the new hidden neuron should be small. If, on the other hand, the sum is small, $v_i$ is not adequately represented by any of these hidden neurons, and 
its weight $w_{ik}$ to the new hidden neuron should be big. 

The algorithm for incremental RBM is summarized in Alg. \ref{alg:callRBM}.

\begin{algorithm}
	\caption{Incremental RBM for a new scene. Initially, there is only one hidden node, i.e., $|\mathbf{h}|=1$, and $t^{iRBM}$ (patience of the model) is set to $\exp(-0.5)$.
		\label{alg:callRBM}}
    \footnotesize
	\KwIn{
		\begin{itemize}
			\item $\textbf{s}$: A new scene (i.e., a $\mathbf{v}$ vector, 
				                s.t. $\mathbf{v}_i=1$ if $\textbf{s}$ contains object with label $i$)
			\item $W$, $|\mathbf{v}|$, $|\mathbf{h}|$: Current model
		\end{itemize}
		}
	\KwOut{$W$: Updated model}
	Clamp $\mathbf{v}$, estimate $\mathbf{h}^0$ and calculate $<v_i h_j>^0$ \Comment{Positive phase}\\
	Reconstruct $\mathbf{v}^1$ from $\mathbf{h}^0$, re-estimate $\mathbf{h}^1$\\
    Calculate $<v_i h_j>^1$   \Comment{Negative phase}\\
    $w_{ij} \leftarrow w_{ij} + \epsilon \times ( <v_ih_j>^0 - <v_ih_j>^1 )$ \Comment{update weights}\\
    $c_v \leftarrow \max\limits_{j} w_{vj}$ \Comment{calculate confidence for visible neurons}\\
    \If{$\exp\left(\min\limits_{v} c_v\right) / Z_0 <  t^{iRBM} \times c^{|\textbf{h}|}_m$} {
       Add a new hidden neuron, let $k$ be its index\\
       $w_{ik} \leftarrow \left(\sum_{j=1}^{|\textbf{h}|-1} w_{ij}\right)^{-1}$ \Comment{Initialize new weights}\\
       $Z_0 \leftarrow \sum_{v} \exp(c_v)$\\
       $c^{|\textbf{h}|}_m \leftarrow \exp\left( \min\limits_{v} c_v \right) / Z_0$\Comment {Update baseline confidence for new $\mathbf{h}$}\\
     }
\end{algorithm}

\subsection{Deep Incremental Boltzmann Machines (diBM)}


If any two contexts in a layer represent similar contextual knowledge, they can be merged and combined under a new context node in an upper layer. To determine whether to add a new hidden layer to the top of the hierarchy, we first define a baseline confidence $r_f$ for a hidden layer $f$ when layer $f$ has exactly two hidden neurons:
\begin{equation}
  r_f \leftarrow d(h_i, h_j), \quad\quad \textrm{for}\ h_i, h_j \in \textbf{h}^f,
\end{equation}
{\noindent}where $d(h_i, h_j)$ is the distance between $h_i$ and $h_j$ based on their weights:
\begin{eqnarray}
d(h_i, h_j) = \frac{1}{2} [ D_{KL}(sm(\textbf{w}^{i}) || sm(\textbf{w}^{j})) +  \\
			D_{KL}(sm(\textbf{w}^{j}) || sm(\textbf{w}^{i}))],
\end{eqnarray}
{\noindent}where $D_{KL}(\cdot||\cdot)$ is the Kullback-Leibler divergence; $\mathbf{w}^{j}=<w_{kj}>$ is the vector of weights connecting $h_j$ to the previous layer's nodes; and $sm(\textbf{w})_i = \exp(w_i) / \sum_j \exp(w_j)$ is the vector-defined softmax.

After adding more neurons to a hidden layer $f$ (i.e., when $|\textbf{h}^f|>2$), the layer's current confidence ($r_f^{curr}\leftarrow \min\limits_{h_i,h_j \in \mathbf{h}^f} d(h_i, h_j)$) drifts away. When that happens, we add a new hidden layer as the next layer of layer $f$. This condition can be defined as follows:
\begin{equation}
 r_f^{curr} < t^{diBM} \times r_f,
\end{equation}
{\noindent}where $t^{diBM}$ is a constant controlling the system's patience to add a new layer with a single neuron. Each neuron in layer $f$ is connected to the single neuron in layer $f+1$ with random weights.

The algorithm for diBM is summarized in Alg. \ref{alg:stackRBM}.

\begin{algorithm}
  \caption{The algorithm for deep incremental BM (diBM). $\mathbb{R}$ initially contains one layer with one hidden neuron. $t^{diBM}$ (patience of the model) is empirically set to $0.1$. \label{alg:stackRBM}}
  \footnotesize
  \KwIn{
       \begin{itemize}
         \item $\textbf{s}$: A new scene (i.e., a $\mathbf{v}$ vector, 
                s.t. $\mathbf{v}_i=1$ if $\textbf{s}$ contains object with label $i$)
         \item $\mathbb{R}=\{\textbf{R}^0, ..., \textbf{R}^l\}$: The current (latent) hierarchy, with $\textbf{R}^i=\{\mathbf{h}^i, \mathbf{W}^i\}$ being the hidden neurons and the weights of layer $i$.
        \end{itemize}
      }
  \KwOut{$\mathbb{R}$: The updated hierarchy.}
  Update each $\textbf{R}^i \in \mathbb{R}$ using Alg. \ref{alg:callRBM}, adding new hidden neurons if necessary\\
  Let $\textbf{R}^f$ be the last layer, and $\mathbf{h}^f$ be its hidden neurons\\
  If $|\mathbf{h}^f| < 2$, set the last layer's baseline confidence, $r_f$, to $0$.\\
  \uIf{Hidden neurons in $\textbf{R}^f$ is incremented, and $|\mathbf{h}^f| = 2$}{
       $r_f \leftarrow d(h_i, h_j)$, for $h_i,h_j\in\mathbf{h}^f$\\
   }
   \uElseIf{$\left[\min\limits_{h_i,h_j \in \mathbf{h}^f} d(h_i, h_j)\right] < (t^{diBM} \times r_f) $}
   {
    $\textbf{R}^{f+1} \leftarrow $ a new incremental RBM layer with one node\\
    $\mathbb{R} \leftarrow \mathbb{R} \oplus \textbf{R}^{f+1}$\Comment{Add new layer to diBM}\\
    $r_f \leftarrow 0$\\
   }
\end{algorithm}

\subsection{Stacked Incremental Restricted Boltzmann Machines (siRBM)}

Another way to construct a hierarchy is by stacking iRBM layers. The number of layers in the stack is determined by comparing confidence of the final iRBM layer (i.e., $\min\limits_{h_i, h_j \in \mathbf{h}} d(h_i, h_j)$) with that of the previous one; if the confidence of the last layer is less, then a new layer is created for stacking. In contrast to the diBM model, siRBM determines adding a new iRBM layer to the hierarchy by assuming encountered scenes are finished for the previous layers.

 \section{Experiments and Results}

In our experiments, we compare iRBM and diBM against (vanilla) RBM (with the same number of hidden units that was found by iRBM), stacked RBM (with the same number of layers and hidden units as found by siRBM), DBM (with the same number of layers and hidden units as found by diBM), incremental RBM \cite{yu2015incremental}, and incremental LDA \cite{CelikkanatContext2014}. In comparing the methods, we use the same number of epochs for each method.

Note that RBM and DBM are not incremental methods; we test them in batch mode (giving all training data at once) and online (incremental mode where we give scenes one at a time). Moreover, we also test how good our diBM can initialize a vanilla DBM method (shown with DBM $\leftarrow$ diBM in the tables) for the tasks used in the paper.

\subsection{Dataset}

For training and evaluating the methods, we used the SUN RGB-D scene classification and segmentation dataset \cite{song2015sun}, which is composed of labeled objects in various scenes. We selected 10,335 scenes by splitting the dataset into two for training (7,000) and testing (3,335). 

\subsection{Quantitative Evaluation of the Number of Contexts Found}

\begin{figure}
\centerline{
\subfigure[Number of context vs encountered scenes]{
	\includegraphics[width=0.5\textwidth]{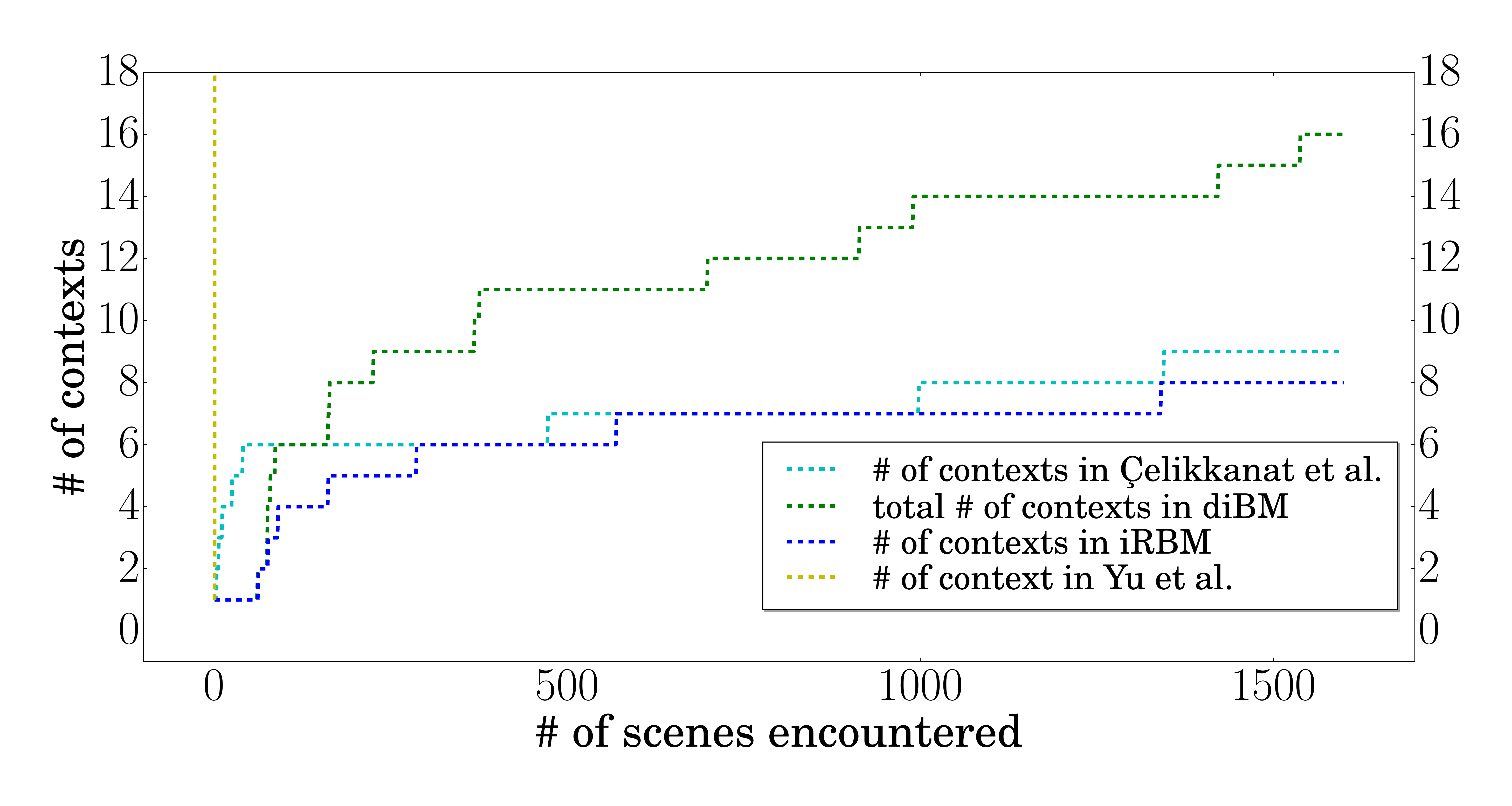}
	}
    }
    \centerline{
\subfigure[Number of hidden layers vs encountered scenes]{
	\includegraphics[width=0.48\textwidth]{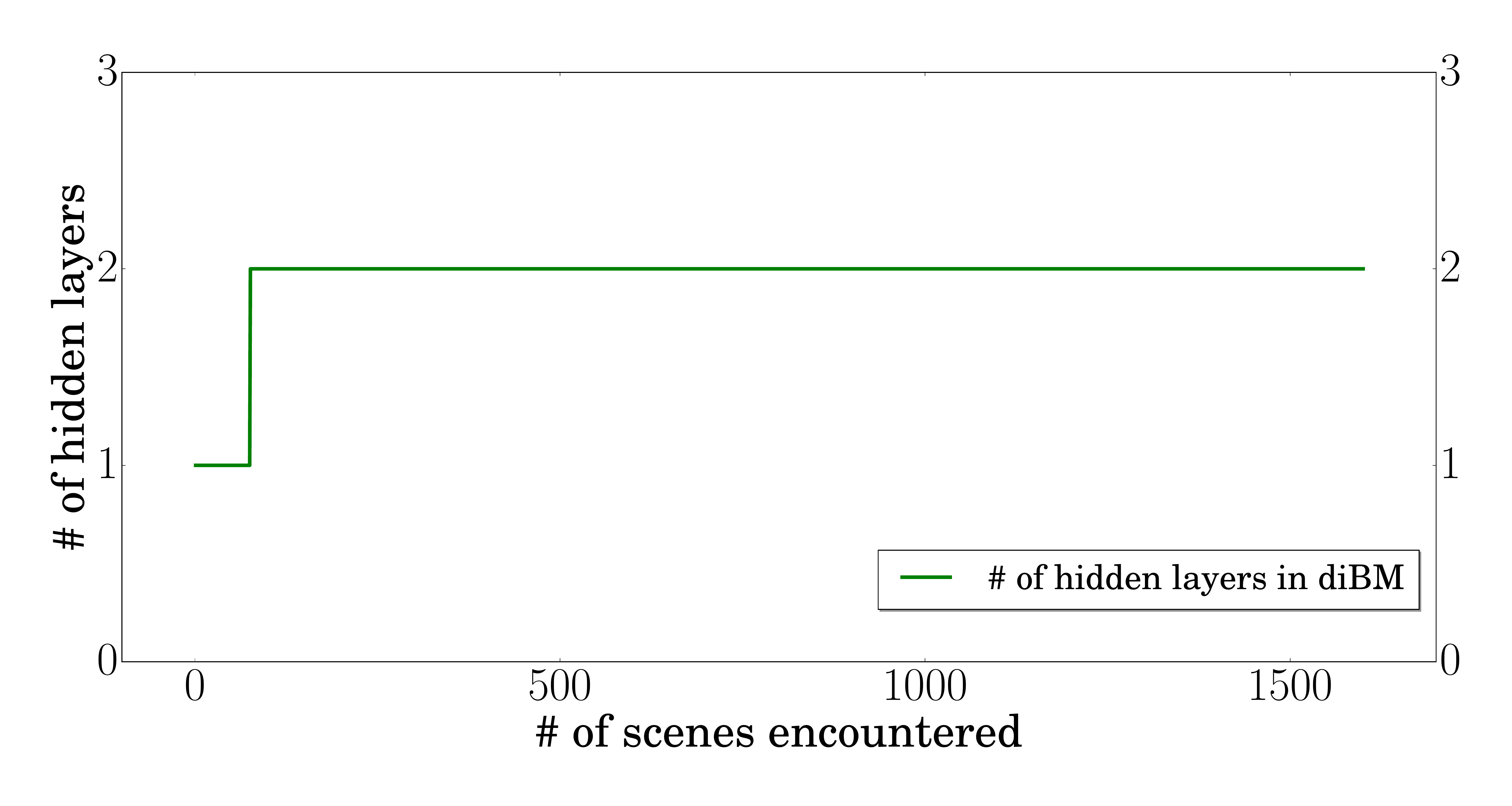}
	}    
}
\caption{Number of hidden layers and topics obtained from SUN RGB-D Dataset with equally distributed 8 contexts and 200 scenes from each context with online learning. Note that the number of hidden layers matters only for diBM. diBM yields a hierarchy with 16 contexts (9 in the first layer, 7 in the second) in total.\label{fig:numtopic2}}
\end{figure}

We first analyze how many contexts and layers are discovered by the incremental methods and diBM. As shown in Fig. \ref{fig:numtopic2}, where the correct number of contexts (scene categories) is 8. From the figure, we see that iRBM finds the correct number of contexts in Fig. \ref{fig:numtopic2}. Since the figure shows the total number of contexts on all layers, the number of contexts found by diBM, which is 16 (9 for the first layer, 7 for the second layer, is more. 


\subsection{Quantitative Evaluation of the Resulting Model Entropies}

We compared the methods based on how the systems' entropies change over time, where entropy is defined as follows (as in \cite{CelikkanatContext2014}): 
\begin{equation}
\hat{H} = \rho H(o | c) + (1 - \rho )  H(c | s),
\end{equation}
{\noindent}where random variables $o$, $c$ and $s$ denote objects, contexts and scenes respectively; $H(\cdot|\cdot)$ denotes conditional entropy; and $\rho$ is a constant (selected as 0.5) controlling the importance of the two terms. The first term measures the system's confidence in observing certain objects given a context, and the second one promotes context confidence given a scene.

\begin{figure}
\centerline{
\subfigure[Entopy change vs encountered scenes]{
	\includegraphics[width=0.49\textwidth]{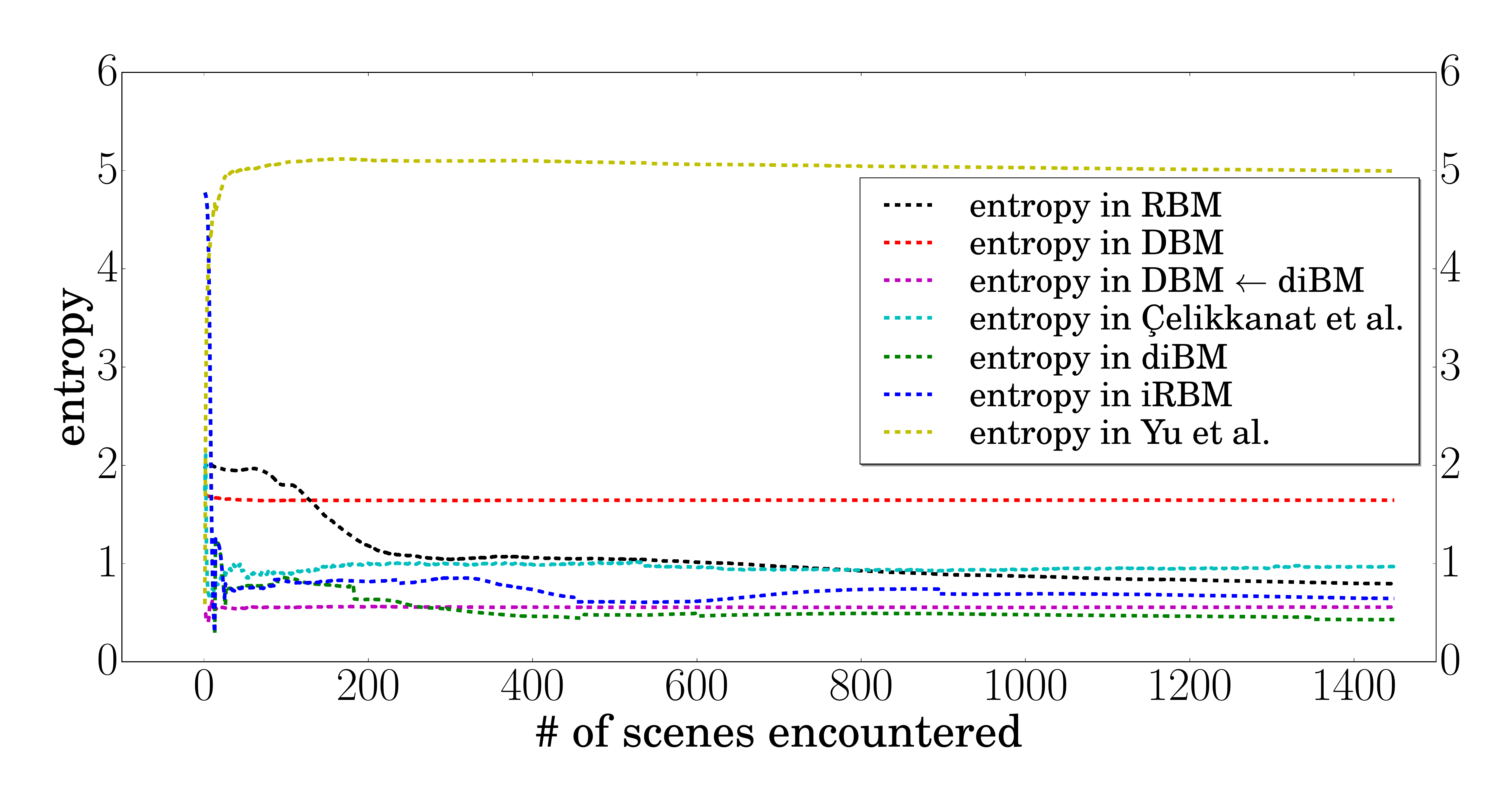}
	}
    }
    \centerline{
\subfigure[Number of context vs encountered scenes]{
	\includegraphics[width=0.49\textwidth]{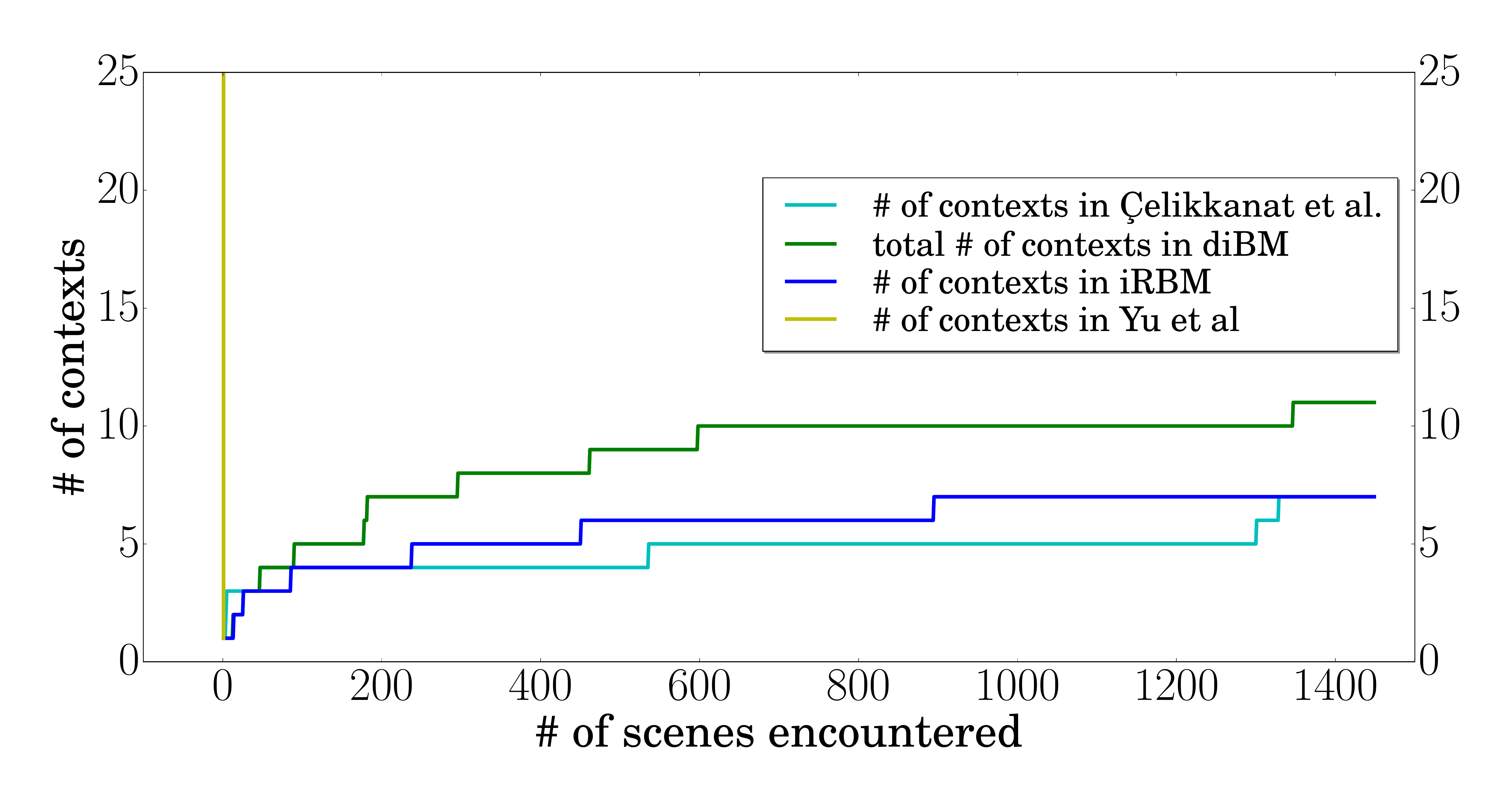}
	}    
}
\caption{Performance of different models on the NYU Depth Dataset. Since the number of hidden neurons is constant in DBM and RBM, they are omitted from the graph where diBM finds 11 contexts in total (8 in the first layer, 3 in the second layer). \label{fig:entropy}}
\end{figure}

Fig. \ref{fig:entropy} displays how the entropies of the incremental models change over time. We see that diBM has discovers a structure with the lowest entropy (we take the mean of its entropy for each layer).

\subsection{Qualitative Inspection of Context Coherence (Hidden Nodes)}

To get a feeling of the performance of the methods, we looked at the strongest objects associated with the hidden neurons. For this, we just compared one-layer methods (iRBM, incremental RBM \cite{yu2015incremental}, incremental LDA \cite{CelikkanatContext2014} and online vanilla RBM) and hence, not considered DBM, diBM, stacked RBM or stacked iRBM, since the first layers of these methods (RBM and iRBM) are included in the comparison. 

Table \ref{tbl:objects_in_context} lists the three best (selected by visual inspection) hidden neurons' strongly connected objects for the different methods. We see that, among the methods, iRBM seems to have found the most relevant objects together in separate contexts. The third hidden neuron of Celikkanat et al. \cite{CelikkanatContext2014} seems to have combined unrelated objects together, and incremental RBM \cite{yu2015incremental} and online vanilla RBM yielded worst results.

\begin{table}
 \caption{Most probable 10 objects of best 3 hidden units of a subset of SUN RGB-D dataset (8 contexts and 1600 scenes). ``d]'' is indeed a label in the dataset. We do not provide results for DBM, Stacked RBM, Stacked iRBM or diBM since they yield similar results for the first layer, when compared to their single layer counterparts, i.e., RBM and iRBM. (We shortened some words to save space: CM: computer monitor, TPD: Toilet Paper Dispenser, ED: Electrical Device) \label{tbl:objects_in_context}}
    \centering
    \begin{tabular}{| c | l | l | l |}\hline
    \multirow{10}{*}{\rotatebox[origin=c]{90}{\parbox[t]{2.25cm}{\centering iRBM \\ (9 contexts found)}}}
    & Hidden1  & Hidden2 & Hidden3   \\ \hline
    & keyboard & oven & sink   \\
    & mouse & stove & toilet \\ 
    & CM & carpet & faucet \\ 
    & cord & countertop & pipe   \\ 
    & chair & toaster & soap  \\
    & cpu & microwave & tap   \\
    & monitor & tilefloor & cabinets  \\ 
    & pillar & refrigerator & urinal\\ 
    & desktop & painting & towel \\ 
    & scanner & plates & TPD \\ \hline
    
    \multirow{10}{*}{\rotatebox[origin=c]{90}{\parbox[t]{2.25cm}{\centering \c{C}elikkanat et al. \cite{CelikkanatContext2014} \\ (9 contexts found)}}}
    & chair & wall & wall \\ 
    & table & keyboard & floor \\ 
    & floor & monitor & sink \\ 
    & wall & computer & toilet \\ 
    & desk & desk & cabinet \\ 
    & door & paper & \textcolor{red}{counter}   \\ 
    & window & mouse & pipe \\ 
    & board & floor & door \\ 
    & bookshelf & door & towel \\ 
    & chairs & window & \textcolor{red}{microwave} \\ \hline
    
    \multirow{10}{*}{\rotatebox[origin=c]{90}{\parbox[t]{2.25cm}{\centering Yu et al. \cite{yu2015incremental} \\ (3527 contexts found)}}}
    & urinal & keyboard & floor   \\ 
    & toilet & book & chair   \\ 
    & towel & monitor & cup    \\ 
    & pipe & pillow & minifridge  \\ 
    & \textcolor{red}{book} & flowers & lid   \\ 
    & sink & mirror & refrigeration   \\
    & window & window & cabinet   \\
    & \textcolor{red}{bookshelf} & adapter & insulatedbag   \\
    & garbage & wall & \textcolor{red}{stallsreflection}   \\
    & \textcolor{red}{bottiles} & floor & frame \\ \hline 
     
    \multirow{10}{*}{\rotatebox[origin=c]{90}{\parbox[t]{2.25cm}{\centering RBM [online] \\ (8 contexts given)}}}
    & mirror & counter & table \\ 
    & sink & teapot & chairs  \\ 
    &  floor & toaster & bookcase \\ 
    & window & coffeemaker & sofa \\ 
    & plumbing & wall & chairline \\ 
    & mop & \textcolor{red}{carrier} & ED \\ 
    & towel & stove & triangle \\ 
    & wall & light & classplate \\ 
    & \textcolor{red}{counter} & \textcolor{red}{d]} & \textcolor{red}{dress} \\ 
    & toilet & cupboard & \textcolor{red}{glass} \\ \hline
\end{tabular}
\end{table}

\subsection{Quantitative Evaluation of Partially Damaged Scene Reconstruction Performance}

Robustness to input noise or noisy estimations is extremely important for any robotic skill. One advantage of contextual information is to provide robustness in these situations by providing a framework in which correctly perceived elements can help prevent erroneous ones. As an example application, we evaluate a scene reconstruction scenario, in which a part of the scene is initially misunderstood, and then corrected via contextual reasoning. For this end, we generated partially-corrupted $\tilde{\mathbf{v}}$ from $\mathbf{v}\in\mathbf{V}$, and we compared methods' reconstruction $\mathbf{v}'$ of the visible vector. For corrupting the visible nodes, we selected $\alpha$ dimensions in $\mathbf{v}$ arbitrarily and flipped those dimensions with probability $0.5$.

\begin{figure}
\centerline{
	\includegraphics[width=0.3\textwidth]{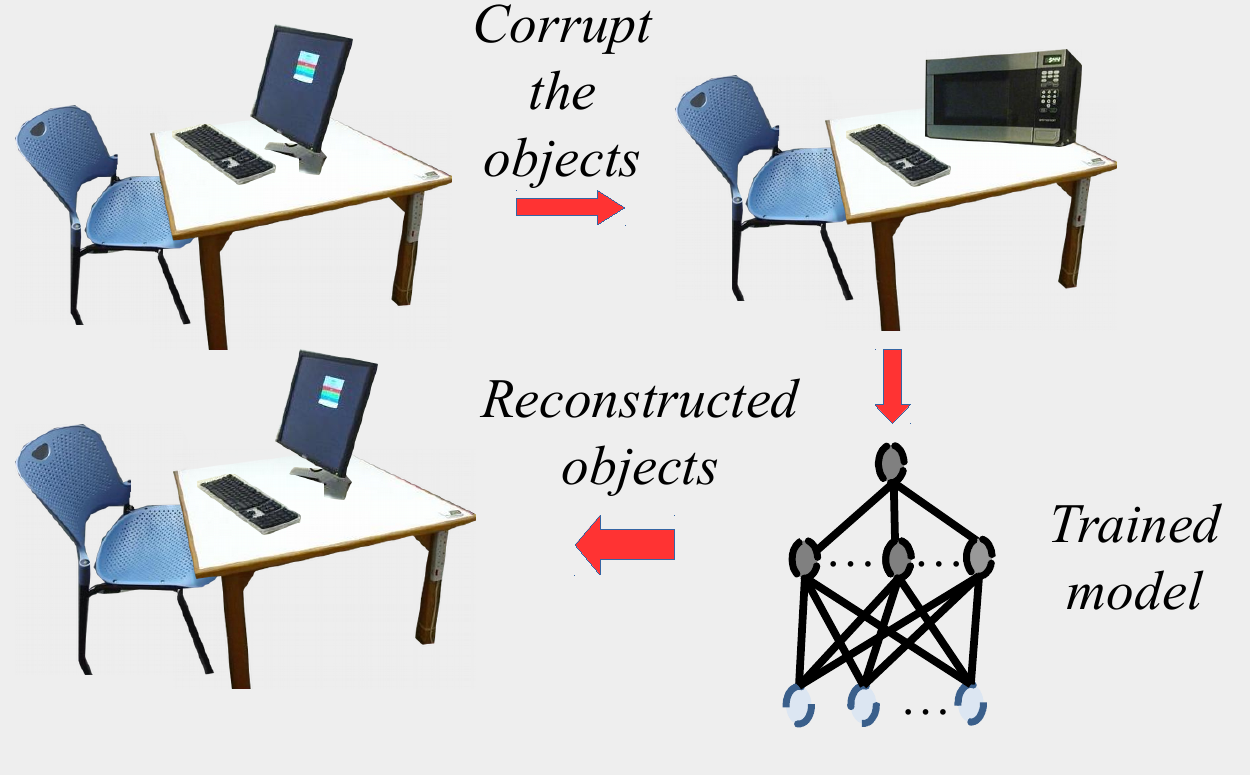}
}
\caption{An illustration of scene reconstruction \label{fig:reconstrut}}
\end{figure}

For evaluating the methods, we devised the following measures:
\begin{eqnarray}\footnotesize
	\textrm{CD} & = & 1 - \frac{\sum_{\mathbf{v}\in \mathbf{V}}\sum_i abs(v_i -\  v_i')}{\alpha|\mathbf{v}|\times |\mathbf{V}|}, \\\footnotesize
	\textrm{CDa} & = & 1 - \frac{\sum_{\mathbf{v}\in \mathbf{V}}\sum_i abs(v_i -\  v_i')}{\sum_{\mathbf{v}\in \mathbf{V}}\sum_i abs(v_i -\  \tilde{v}_i)},
\end{eqnarray}
{\noindent}where CD and CDa respectively are acronyms for Corrupted Dimensions and Corrupted Data; and, $abs(\cdot)$ is the absolute value function. Note that CDa and CD can take negative values if a method destroys more than it successfully reconstructs.

We noticed that, while trying to recover the corrupted bits, some methods destroyed the uncorrupted parts as well. To be able to measure this, we devised alternative versions:
\begin{eqnarray}\footnotesize
	\textrm{CD$^{k}$} & = & 1 - \frac{\sum_{\mathbf{v}\in \mathbf{V}}\sum_i abs(u_i -\  u_i')}{\alpha|\mathbf{v}|\times |\mathbf{V}|}, \\\footnotesize
	\textrm{CDa$^{k}$} & = & 1 - \frac{\sum_{\mathbf{v}\in \mathbf{V}}\sum_i abs(u_i -\  u_i')}{\sum_{\mathbf{v}\in \mathbf{V}}\sum_i abs(v_i -\  \tilde{v}_i)},
\end{eqnarray}
{\noindent}where $\textbf{u}$ is the corrupted part of $\mathbf{v}$.
\begin{table}
 	\caption{Reconstruction performances of the methods for a corruption rate ($\alpha$) of 40\% in the testing subset of the SUN RGB-D dataset \cite{song2015sun}. KCP and UCP stand for Known Corrupted Part and Unknown Corrupted Part, respectively. \label{tbl:acc_sun}}
 	\centering\footnotesize
 	\begin{tabular}{ | c | l | c | c | c | c |}
 		\hline
 		& & \multicolumn{2}{|c|}{KCP} & \multicolumn{2}{|c|}{UCP}\\ 
 		& &    CD$^k$ & CDa$^k$  & CD & CDa \\ \hline \hline
 		\multirow{4}{*}{\rotatebox[origin=c]{90}{\parbox[t]{1cm}{\centering Batch}}} 

 		 & RBM    & 0.724 & 0.449 & 0.304 &  -0.391\\ 
          & Stacked RBM & 0.998 & 0.997 &  0.996 & 0.992 \\ 
          & DBM  & 0.997 & 0.993 &  0.992 &  0.984 \\ 
          & DBM $\leftarrow$ diBM  &   0.993 & 0.985 &  0.982  & 0.964 \\ \hline

		
 		\multirow{4}{*}{\rotatebox[origin=c]{90}{\parbox[t]{2.25cm}{\centering Online}}}
 		& RBM &  0.752 & 0.504 &  0.373 & -0.253  \\ 
         & Stacked RBM  &  0.998 & 0.996 &  0.996 & 0.993 \\ 
         & iRBM &  0.962  & 0.925 &  0.906 &  0.812\\ 
         & Stacked iRBM &  0.997 & 0.995 &  0.994 & 0.987 \\ 
         & diBM   & \textbf{0.999} & \textbf{0.998} &  \textbf{0.997} & \textbf{0.994} \\
         & DBM &  0.997 & 0.994  & 0.993 & 0.989 \\ 
         & DBM $\leftarrow$ diBM     & 0.997 & 0.993   & 0.991 &  0.983\\ 
         & Yu et al. \cite{yu2015incremental}  & 0.521 & 0.042 & -0.099 &  -1.200\\ \hline	
 	\end{tabular}
 \end{table}

		

Table \ref{tbl:acc_sun} lists the accuracies for the different methods. We see that, among the batch methods (that use all data at once), stacked RBM performs best, in fact better surprisingly better than DBM or DBM initialized with diBM weights. This suggests that stacked RBM can converge faster than these methods. When we look at the performances of the incremental methods, they obtain better or on par results compared to the batch ones and that diBM yields the best results, not only better than its incremental competitors but also the batch methods. The main reason for this performance is the fact that we force sparse strong connections between objects and contexts. 

Comparing iRBM with RBM, stacked iRBM with stacked RBM, and diBM with DBM from Table \ref{tbl:acc_sun}, main conclusion is that our methods converge (incrementally) to a model that is assumed to have a given structure. This suggests that, with methods like ours, we can build (evolve) models through time that perform as good as (and in fact, better than) their rigid counterparts. This alleviates the problem of model (structure) selection before or while training models.

See Fig. \ref{fig:reconstrut} for an example corrupted and reconstructed scene.

\section{Conclusion}
We have proposed two new methods in the paper: (i) one method to incrementally construct a layer of RBM by pushing hidden neurons to favor a subset of the objects, and vice versa, and (ii) another method to incrementally add hidden layers with each arriving scene to construct a deep incremental BM. Compared to baseline methods and other methods in the literature, we showed in the SUN RGB-D Dataset that our methods construct better models in learning a distribution of the data, as shown by the correctly found number of contexts, the low entropy, the reconstruction of the corrupted data and visual inspection of what the hidden neurons represent.
 
\section*{Acknowledgment}

This work was supported by the Scientific and Technological Research Council of Turkey (T\"UB\.{I}TAK) through project called ``Context in Robots'' (project no 215E133).

\bibliographystyle{IEEEtran}
\bibliography{references}

\end{document}